\documentclass{article}
\usepackage{ijcai20}

\usepackage{times}
\usepackage{soul}
\usepackage{url}
\usepackage[hidelinks]{hyperref}
\usepackage[utf8]{inputenc}
\usepackage[small]{caption}
\usepackage{graphicx}
\usepackage{amsmath}
\usepackage{amsthm}
\usepackage{booktabs}
\usepackage{algorithm}
\usepackage{algorithmic}
\usepackage{subcaption}
\usepackage{latexsym}
\usepackage{amsmath}
\usepackage{multirow}
\usepackage{tablefootnote}
\usepackage{comment}
\usepackage{amssymb}
\usepackage{url}
\usepackage{enumitem}
\usepackage{array}
\usepackage{diagbox}
\setlist{nosep}

\usepackage{color}

\newcommand{\newcite}[1]{\citeauthor{#1} \shortcite{#1}}
\usepackage{graphicx,calc}
\DeclareMathOperator*{\argmax}{arg\,max}
\newcommand{\eps}{\textit{eps}}
\newcommand{\sampl}{\textit{sampl}}
\newcommand{\agmax}{\textit{argmax}~}

\title{Learn How to Cook a New Recipe in a New House: Using Map Familiarization, Curriculum Learning, and Bandit Feedback to Learn Families of Text-Based Adventure Games}

\author{
Xusen Yin$^1$
\and
Jonathan May$^1$
\affiliations
$^1$Information Sciences Institute/University of Southern California
\emails
\{xusenyin, jonmay\}@isi.edu
}

\begin{document}

\maketitle

\begin{abstract}
We consider the task of learning to play families of text-based computer adventure games, i.e., fully textual environments with a common theme (e.g. cooking) and goal (e.g. prepare a meal from a recipe) but with different specifics; new instances of such games are relatively straightforward for humans to master after a brief exposure to the genre but have been curiously difficult for computer agents to learn. We find that the deep Q-learning strategies that have been successfully leveraged for superhuman performance in single-instance action video games can be applied to learn families of text video games when adopting simple strategies that correlate with human-like learning behavior. Specifically, we build agents that learn to tackle simple scenarios before more complex ones using curriculum learning, that familiarize themselves in an unfamiliar environment by navigating before acting, and that explore uncertain environments more thoroughly using contextual multi-armed bandit decision policies. We demonstrate improved task completion rates over reasonable baselines when evaluating on never-before-seen games of that theme.
\end{abstract}

\section{Introduction}

Building agents able to play text-based adventure games is a useful proxy task for learning open-world goal-oriented problem-solving dialogue agents. Via an alternating sequence of natural language descriptions given by the game and natural language commands given by the player, a player-agent navigates an environment, discovers and interacts with entities, and accomplishes a goal, receiving explicit rewards (a.k.a. scores) for doing so. Human players are skilled at text games when they understand the situation they are placed in and can make rational decisions based on their life and game playing experience. For example, in the classic text game Zork \cite{Zork-paper}, the adventurer discovers an air pump and an uninflated plastic boat; common sense leads human players to \textit{inflate} the boat with the pump; or in an unfamiliar environment, human players try new actions other than relying on their game experience.

Games such as Zork are very complicated and are designed to be played repeatedly until all the puzzles contained within have been solved; in this way, they are not very similar to real human experiences. Another kind of text game, as exemplified by the TextWorld learning environment \cite{textworld-a-learning-environment-for-text-based-games} and competition, expects agents to learn a particular task theme (such as rescuing victims from a burning building or preparing a meal) but evaluates on never-before-seen instances of that theme in a zero-shot evaluation setting. This is a much more realistic scenario. A person who has never cooked a meal before would no doubt flounder when asked to prepare one. In order to learn to cook, one does not begin by learning to make Coq au Vin, but rather starts simply and works up to more complicated tasks. However, once the cooking skill is learned, one would reasonably expect to be able to prepare a new recipe the first time it is seen. Furthermore, even if the recipe was prepared in a somewhat unfamiliar location (say, the kitchen of a vacation home), a reasonable person would explore the new space, recognize the familiar rooms and elements, and then begin cooking. 

In this work, we approach this more-realistic scenario and consider how we might train models to learn to play familiar but unseen text games by adopting a training regimen that mirrors human skill acquisition. We additionally show that, by exploring the search space more thoroughly and evenly by leveraging multi-armed bandit feedback, an agent can reach higher scores in the zero-shot evaluation setting. Specifically, we make the following contributions in our text game agent learning  models:
\begin{itemize}
    \item We build agents that can play unseen text-based games. We show how the proper use of domain-aware curriculum learning strategies can lead to a better learned agent than learning with all games at once.
    \item We draw a distinction between knowledge into the \textit{universal} (e.g., that cooking can be done in the kitchen) and \textit{instance} (e.g. that the kitchen is east of the bedroom); the former can be usefully learned with training data, but the latter cannot. We show how \textit{environment familiarization} through construction of a knowledge graph improves learning. 
    \item We show that incorporating bandit feedback during evaluation leads to a better agent by exploring environments more thoroughly, especially in a zero-shot test with new games.
\end{itemize}


\section{Reinforcement Learning for Text Games}
\label{sec:rl}

The influential Deep Q-Network (DQN) approach of learning simple action video games pioneered by \newcite{google-atari} has motivated research into the limits of this technique when applied to other kinds of games. We follow recent work that ports this approach to \textit{text-based games} \cite{D15-1001,P16-1153,fulda2017affordance,DBLP:conf/nips/ZahavyHMMM18,DBLP:journals/corr/abs-1805-07274,DBLP:conf/cig/KostkaKKR17,DBLP:journals/corr/abs-1806-11525,DBLP:journals/corr/abs-1812-01628,my-own}. The core approach of DQN as described by \newcite{google-atari} is to build a \textit{replay memory} of partial games with associated scores, and use this to learn a function $f_{DQN}: (S, A) \to \mathcal{R}$, where $f_{DQN}(s,a)$ predicts a Q-value, which is the future discounted {\it return} until the game terminates, obtained by choosing action $a \in A$ when in state $s \in S$. From $s$, choosing $\argmax_{a \in A} f_{DQN}(s,a)$ affords the optimal action policy.
 
\begin{figure}[h]
\centering
    \includegraphics[width=0.4\textwidth]{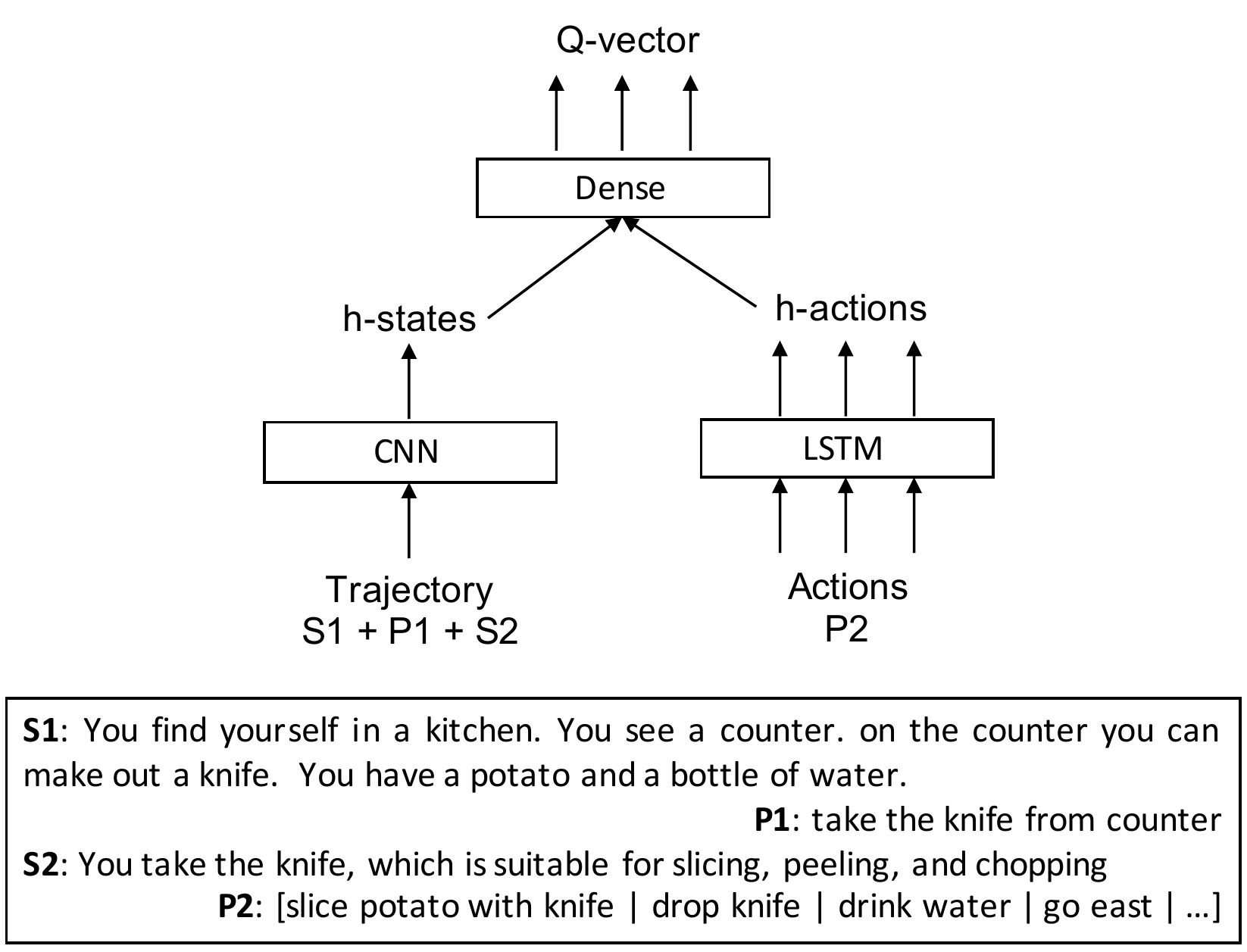}
    \caption{The architecture of the DRRN model. Trajectories and actions are encoded by a CNN and an LSTM into hidden states and hidden actions, respectively, followed by a dense layer to compute the Q-vector. On the bottom, we show a truncated example of dialogue from a text game in the cooking genre, with S1 and S2 representing system description, and P1 showing the player's first actions towards S1. S1 + P1 + S2 is an example of a trajectory. P2 shows a set of actions to choose from.}
   \label{fig:archi-drrn}
\end{figure}

As in the original work, a key innovation is using the appropriate input to determine the game state; for video games, it is using a sequence of images (e.g. 4-frame of images in \cite{google-atari}) from the game display; while for text games we use a history of system description-player action sequences, which we call a \textit{trajectory}; an abbreviated example is given in Figure \ref{fig:archi-drrn}. A means of efficiently representing infinite $S$ is necessary; most related work uses LSTMs \cite{D15-1001,DBLP:journals/corr/abs-1812-01628,DBLP:journals/corr/abs-1806-11525,DBLP:conf/cig/KostkaKKR17,DBLP:journals/corr/abs-1805-07274}, though we follow \cite{DBLP:conf/nips/ZahavyHMMM18,my-own}, which uses CNNs, to achieve greater speed in training. The DQN is trained in an exploration-exploitation method ($\epsilon$-greedy): with  probability $\epsilon$, the agent chooses a random action (explores), and otherwise the agent chooses the action that maximizes the DQN function. The hyperparameter $\epsilon$ usually decays from 1 to 0 during the training process. At inference time, $\epsilon=0$ is often used to choose the optimal action from the policy.

 Much game-learning research is concerned with the optimization of a \textit{single game}, e.g. applying DQN repeatedly on Pac-Man with the goal of learning to be very good at playing Pac-Man. While this is a realistic goal when strictly limited to the domain of video game play,\footnote{occasional stochasticity notwithstanding} single-game optimization is rather unsatisfying. It is difficult to tell if a single game-trained model has managed to simply overfit on its target or if it has learned something general about the task it is trying to complete. More concretely, if we consider game playing as a proxy for real-world navigation (in the action game genre) or task-oriented dialogue (in the text genre), it is clear that a properly trained agent should be able to succeed in a new, yet familiar environment. We thus depart from the single-game approach taken by others \cite{D15-1001,P16-1153,DBLP:journals/corr/abs-1812-01628,DBLP:conf/nips/ZahavyHMMM18} and evaluate principally on games that are in the same genre as those seen in training, but that have not previously been played during training.
 
\subsection{Handling Unbounded Action Representations}
 
A consequence of learning to play a game that has not been seen before is that actions not seen in training may be necessary at test time. Vanilla DQNs as introduced by \newcite{google-atari} are incompatible with this modification; they presume a predefined finite action space and were tested for a space of up to 18 (each of nine joystick directions and a potential button push). Additionally, vanilla DQNs presume no semantic relatedness among action spaces, while in text games it would make sense for, e.g., \textit{open the door} to be semantically closer to \textit{shut the door} than \textit{dice the carrot}. In our experiments we assume a game's action set is fully known at inference time but not beforehand, and that actions have some relatedness.\footnote{This is itself still a simplification, as many text games allow open text generation and thus infinite action space. Our approach does not preclude abandoning this simplification, but the difficulty of the problem is sufficient to leave this for future work.} We thus represent actions using  Deep Reinforcement Relevance Networks (DRRN) \cite{P16-1153} (Figure \ref{fig:archi-drrn}), a modification of the standard DQN. Actions are encoded via an LSTM \cite{lstm-original} and scored against state representations for Q-values according to this equation: $f_{DRRN}(s, a) =  h_s W h_a$, where $W$ is a learned weight matrix, $h_s$ is the encoded state and $h_a$ is the encoded action. In preliminary experiments we found that LSTMs worked better than CNNs on the small and similar actions in our space such as \textit{take yellow potato from fridge} and \textit{dice purple potato}. In practice, we compute the Q-vector for each state at once with all actions $A$ as $f_{DRRN}(s, A)$, making the DRRN as efficient as a normal DQN. 

\section{Methods}

A {\it direct} application of DQN-like algorithms on families of games may cause problems. Though there has been some work \cite{google-atari,DBLP:journals/corr/abs-1805-07274} learning to play multiple games with single DQN models, these work deviate from our situation: families of games (over thousands) training and zero-shot evaluation. We consider three problems that arise in our scenario, 1) how to arrange the training process with over thousands games; 2) how to learn possible contradicted knowledge from game to game; and 3) how to utilize the inadequately learned policies for zero-shot evaluation. We propose three methods to alleviate these problems.

\subsection{Curriculum Learning}
\label{sec:curric}

Correctly training a DQN-like model to play even a single game can take millions of training steps \cite{google-atari,my-own} due to the need for heavy exploration. If our models are able to learn critical general skills in the early parts of training, they can focus on more fine-grained skills later on. For example, recognizing that the action \textit{cook potato with stove} matches the cookbook instruction \textit{fry potato} allows generalization to, e.g., \textit{fry eggplant}. This skill is needed across all games. More specific skills, like knowing to \textit{drop} items before \textit{picking up} other items are less commonly used. 
Curriculum learning \cite{Bengio:2009:CL:1553374.1553380} is a good way of structuring our learning to capture core skills first and gradually build in more complicated knowledge. For families of games with natural levels of difficulty, we can arrange games into difficulty levels, and train an agent with simple games first.

\subsection{Learning Universally from Local Information}
\label{sec:go-strategy}

Since knowledge like the connection between the behavior of {\it fry} and using a {\it stove} can be learned from past experience and applied to future scenarios, we call this {\bf universal knowledge}. Other knowledge that is specific to a particular scenario and not reusable we term {\bf instance knowledge}. In a specific game, for example, the player may have to \textit{go north} to reach the \textit{kitchen}. However, this will not be the case in general. Thus, naively learning a policy for the action \textit{go east} given a particular state is likely to be sub-optimal. We'd like to ensure that training does not overfit for each single game by turning instance knowledge into universal knowledge.

As it turns out, in the domain we are studying, learning that we must \textit{go} from the room we are in (generally to reach the kitchen or a room containing missing ingredients) \textit{is} universal knowledge. A simple way to turn instance into universal knowledge, which we call \textit{random-go}, is to conflate all actions of the form \textit{go \texttt{direction}} into a single \textit{go} action when choosing actions from learned policy, but then randomly choose a cardinal direction.

Since the room we are trying to reach is more universally important than the direction chosen in a particular game, another approach to converting instance to universal knowledge is to augment directions with the name of the room that will be reached before encoding actions. If, in a particular game, the \textit{bedroom} is \textit{east} of the \textit{hallway}, the action \textit{go east} is modified to be \textit{go east to hallway}, enabling the action representation to incorporate the more globally useful room type of context into its representation.

\subsection{Zero-shot Evaluation with Uncertainty}
\label{sec:uncertainty}

Using the learned policy in a greedy way by choosing actions with \agmax (Section \ref{sec:rl}) leaves no possibility of randomization.
Since the policy is learned by the function $h_s W h_a$, underestimated representations for infrequently seen state-action pairs may contain high variance, which leads to the selection of incorrect Q-values and hence sub-optimal policies, leading to repetition of choosing bad actions \cite{my-own}.
The phenomenon becomes more severe in the setting of zero-shot evaluation with unseen games, especially when encoding long unseen trajectories.

To mitigate the problem, instead of using the learned policy greedily, two commonly used variants during evaluation to add stochasticity are to 1) use a fixed small $\epsilon$, e.g. 0.05, to allow additional exploration \cite{D15-1001,my-own,DBLP:conf/nips/ZahavyHMMM18,DBLP:journals/corr/abs-1806-11525} and 2) sample actions from the policy according to action distributions \cite{P16-1153}.
However, actions that are randomly chosen can have dire results. For example, making the decision to cook an ingredient that has already been cooked will result in destruction of that ingredient and a game failure. Also, when playing unseen games, we should give more opportunity to those state-action pairs that are less seen in the training games, i.e. exploring more in a new environment.

We instead model the uncertainty of choosing actions by employing contextual multi-armed bandit feedback.
We assume that during evaluation the Q-value for each action $a$ is linearly related with the encoded state $h_s$,
w.r.t. the coefficient $\theta_a$,
plus noise $\eta_a$ drawn from an R-sub-Gaussian with a covariance $V_a$: $Q_{s,a}=h_s^T\theta_a+\eta_a$.
We can solve for parameters $\theta_a$ with ridge regression loss $\|Q_{s, a} - h_s^T\theta_a\|_2^2 + \lambda\|\theta_a\|_2^2$,
where $\lambda$ is the hyperparameter to control the size of $\theta_a$.
With ridge regression, we derive the covariance matrix $V_a$ at step $t$ of one game play:
\[V_{a}^{t} = \lambda I + \sum_{j: a^j=a} h_{s}^{j}(h_{s}^{j})^T.\]

For the next step $t+1$ of game playing, we compute a confidence bound for $Q_{s, a}^{t+1}$ as $c_t \sqrt{(h_{s}^{t+1})^T (V_{a}^{t})^{-1}h_{s}^{t+1}}$, where $h_s^{t+1}$ is the next step's hidden state and $c_t$ is a normalization value related to $V_a^t$, as detailed in \cite{DBLP:conf/nips/ZahavyHMMM18,NIPS2011_4417}. Then we select an action that has the largest upper bound of $Q_{s, a}^{t+1}$:
\[a^{t+1} = \argmax_{a\in A} \left(Q_{s, a}^{t+1} + c_t \sqrt{(h_{s}^{t+1})^T(V_{a}^{t})^{-1}h_{s}^{t+1}}\right).\]

The Q-value with upper bound solved with the linear relation assumption is known as Linear Upper Confidence Bound (LinUCB) \cite{Auer2003,Abe2003RL,NIPS2011_4417}. 
The intuition behind this is that rarely seen state-action pairs should be explored more, so that we may have more confidence in the $Q$ value of this (state, action) pair.
 The LinUCB method is not commonly used in the training of DQN-like models, since it requires the feature space for context encoding ($h_s$ in our scenario)
to not change during training. This is  inherently contradicted by the DQN framework. Although  \newcite{DBLP:conf/nips/ZahavyHMMM18} use LinUCB to estimate
the elimination signal of actions along with DQN training, they utilize a batch-update training framework that updates $h_s$ in a batch, as opposed to in every training step. We instead use LinUCB at \textit{evaluation} time, when the feature space is fixed. 

\section{Experiment Setup}

\subsection{Games}
\label{sec:games}

We use the games released by Microsoft for the `First TextWorld Problems' competition. The competition provides 4,440 cooking games generated by the TextWorld framework \cite{textworld-a-learning-environment-for-text-based-games}. The goal of each game is to prepare a recipe. The action space is simple, yet expressive, and has a fairly large, though domain-limited, vocabulary. 
A portion of a simple example is shown in Figure~\ref{fig:archi-drrn}.

The games are divided into 222 different \textit{types}, with 20 games per type. A type is a set of attributes that increase the complexity of a game. These attributes include the number of ingredients, the set of necessary actions, and the number of rooms in the environment. One example of such a type is {\it recipe3 + take3 + open + drop + go9} that implies the game contains three ingredients in the recipe, and players need to find and take the three items. In the process of finding these items, there could be doors to open, e.g. a door of a fridge, or a door of a room. The agent may also need to drop an item it is holding before taking another. Finally, the {\it go9} means there are nine different rooms in the game. A constant reward (i.e. one point) is given for each acquisition or proper preparation of a necessary ingredient as well as for accomplishing the goal (preparing the correct recipe), according to the game design. Each game has a different maximum score, so we report aggregate scores as a {\bf percentage of achievable points} for evaluation results. Admissible actions are available upon request in the TextWorld cooking games at each step.

\subsection{Game Preparation}

We divide the game types into six \textit{tiers} of increasing difficulty. The easiest games take place inside a single room and require only one (tier-1), two (tier-2), or three (tier-3) ingredients. More complicated are the multi-room games; these may have six (tier-4), nine (tier-5), or twelve (tier-6) rooms. Intuitively, it should be very easy to learn a tier-1 game. Adding additional ingredients requires knowing how to prepare each ingredient correctly, and adding additional rooms requires finding the kitchen and other locations. Table~\ref{tbl:tiers} contains per-tier details.

\begin{table}[]
\centering
\begin{tabular}{rrrrr}
tier & \#ingredients  & \#rooms & \#games & max scores \\\hline
1    & 1         & 1      & 420  &  1317 \\
2    & 2        & 1      & 420   &  1891 \\
3    & 3       & 1      & 420    &  2440 \\
4    & $\le$ 3        & 6     & 1040    & 4120 \\
5    & $\le$ 3       & 9     & 1040  &   4028 \\
6    & $\le$ 3       & 12     & 1040  & 4110 \\ 
\end{tabular}
\caption{Tiers of games. The tiers are selected by the difficulty level of games. Tier-1 is the simplest, containing only one ingredient in a recipe and one room to explore per game. Tier-6 is the most difficult, including up to three ingredients in a recipe, and twelve rooms to explore per game. The first three tiers only contain one room, which means there need be no {\it go} actions involved in these games. Maximum scores summed up over each tier are shown in the last column.}
\label{tbl:tiers}
\end{table}

We hold out a selection of 10\% of the games and divide this portion into two separate test sets, each consisting of 222 games, one from each type. We randomly select an additional 400 games as a dev set and keep the remaining games for training. We consider an \textit{episode} to be a play-through of a game; there are multiple episodes of each game run during training and scores are taken over a 10-episode run of each game when evaluating test. An episode is run until a loss (an ingredient is damaged or the maximum of 100 steps is reached) or a win, by completing the recipe successfully. Apart from the inherent game reward, we add $-0.1$ reward (i.e. punishment) to every step, to encourage more direct gameplay. Also, if the game stops early because of a loss, we set the instant reward to $-1$ to penalize the last dire action. We use the reward shaping methods only for model training, and we report raw scores earned from games without any modification during evaluation.

\subsection{Hyperparameters}
During training, we use 50,000 observation steps and 500,000 replay memory entries. From a training run, we select the model with the highest score on the dev set for test inference. The maximum total steps of evaluating on one test set is thus $222\times 10\times 100$. The maximum total score is not unique since different games could have different scores (Table \ref{tbl:tiers}). We use the percentage of scores as the evaluation criteria in the following sections. The higher the score, the better the agent. We also show a detailed \textit{percentage} of scores for each tier. The average of scores of all tiers may not be equal to the all-tier result, because the total maximum scores for each tier may differ. Unless otherwise noted (e.g. in Section \ref{sec:uncertainty-experiment}), we use the greedy method with \agmax to choose the best action from the learned policy during evaluation.

We use a CNN with 32 of each size-3, 4, 5 convolutional filters, followed by a max-pooling layer, following the encoder architecture of \cite{my-own}. The LSTM action encoder contains 32 units in a single layer. We use the last LSTM hidden state as the encoded action state.
We initialize our models with random word embeddings and position embeddings. We use a fixed embedding size of 64. At every training step, we draw a minibatch of 32 samples and use a learning rate of $1e-5$ with the Adam optimizer \cite{DBLP:journals/corr/KingmaB14}. We trim trajectories to contain no more than 11 sentences and 1,000 tokens to avoid unnecessarily long concatenated strings.

\section{Experiments and Discussion}
\subsection{Core Results}

\begin{table}[h]
\centering
\begin{tabular}{l|l|rr}
\multirow{2}{*}{} & \multirow{2}{*}{Experiment} & \multicolumn{2}{c}{Score \%} \\
                         &   & Test 1 & Test 2 \\
\hline\hline
baseline & random & 14 & 14 \\\hline
\multirow{2}{*}{\shortstack[l]{curriculum\\(Section \ref{sec:curric})}}& mixed  & 50 & 54 \\
& curric & 64 & 64 \\\hline
\multirow{3}{*}{\shortstack[l]{familiarization\\(Section \ref{sec:go-strategy})}} & go-cardinal & 50 & 52 \\
& go-random   & 55 & 57 \\
& go-room & 55 & 58 \\\hline
\multirow{4}{*}{\shortstack[l]{uncertainty\\(Section \ref{sec:uncertainty})}} & $\epsilon=0$ & 64 & 64 \\
& $\epsilon=0.05$ & 63 & 63 \\
& sampling & 69 & 67 \\
& LinUCB & 72 & 68 \\
\end{tabular}
\caption{Core overall results on unseen games of all difficulty levels. The random action baseline gives predictably poor results. Using curriculum learning (curric) is preferred to training with all games simultaneously (mixed). Casting directions in terms of the room destination (go-room) generalizes better than learning specific cardinal directions (go-cardinal), but the alternative of picking a direction at random (go-random) appears surprisingly competitive. Evaluation with LinUCB can further improve scores by more thorough exploration, comparing with greedy policy usage $\epsilon=0$, $\epsilon=0.05$ to allow stochasticity, and sampling from policy distribution.}
\label{tab:core}
\end{table}

Core findings are shown in Table~\ref{tab:core}. For a simple, training-free baseline, we choose a random action from the set of admissible actions at each state. Our main comparisons are that of curriculum learning (curric) as described in Section~\ref{sec:curric} to the default (mixed), and between the three different approaches to handling instance knowledge as described in Section~\ref{sec:go-strategy}, and evaluation that compares the use of LinUCB to the greedy method, a  small $\epsilon$ to allow some inference-time stochasticity, and sampling from policy distribution as described in Section~\ref{sec:uncertainty}. 
We next take a more in-depth look at the differences in learning behavior.

\subsection{Curriculum Analysis}
\label{sec:curric-analysis}


\begin{table}[!htb]
\centering
\begin{tabular}{lrrrr}
\multirow{2}{*}{Tier}    & \multicolumn{2}{c}{Test 1} & \multicolumn{2}{c}{Test 2} \\
    & mixed & curric & mixed & curric \\ \hline
1   & 88  & {\bf 96}   & 85  & {\bf 100} \\ 
2   & 53  & {\bf 75}   & 53  & {\bf 70} \\ 
3   & 57  & {\bf 61}   & 54  & {\bf 71} \\ 
4   & 55  & {\bf 68}   & 57  & {\bf 69} \\ 
5   & 40  & {\bf 64}   & 55  & {\bf 63} \\ 
6   & 36  & {\bf 46}   & 41  & {\bf 43} \\ \hline
All & 50  & {\bf 64}   & 54  & {\bf 64} \\ 
\end{tabular}
\caption{Comparing the evaluation results of training all tiers together (mixed) and training with curriculum learning (curric) on the two separate test sets. Rows 1-6 show the breakdown of total scores and steps on each tier. The curriculum learning method generally shows better results on both test sets.}
\label{tbl:all-vs-curriculum}
\end{table}

\begin{figure}[h]
    \centering
    \includegraphics[width=0.35\textwidth]{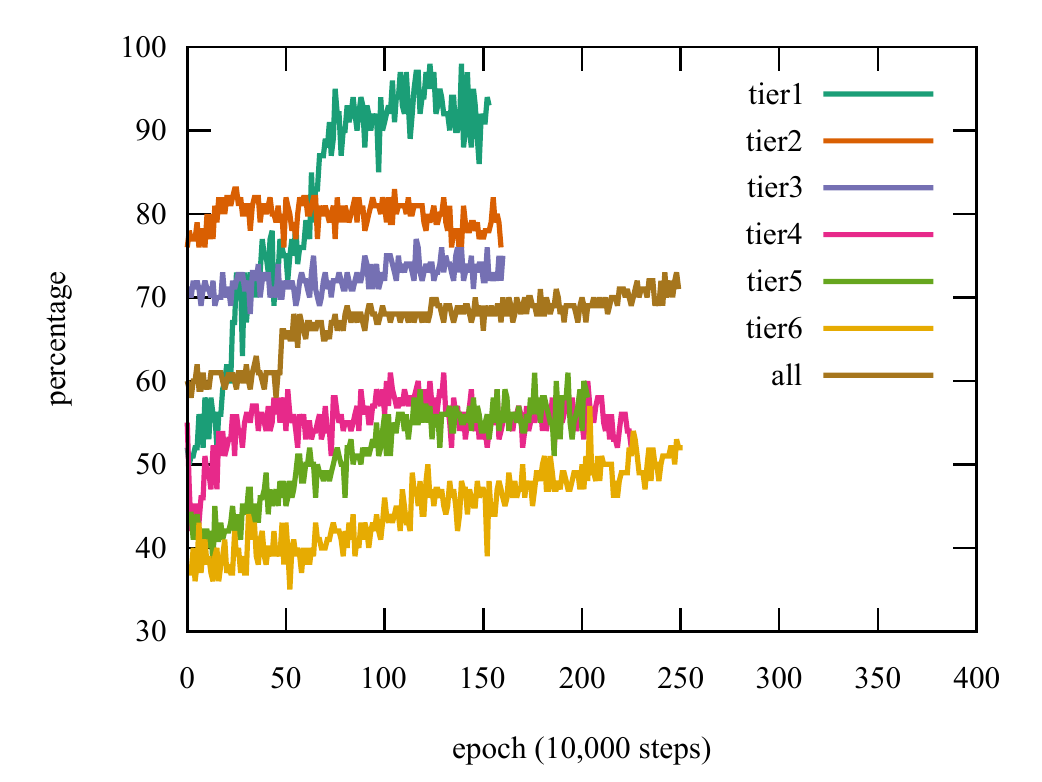}
     \caption{The training process of `curric' with go-room strategy broken down by tier. Fine tuning over all tiers is marked as `all'. Results on tier-specific dev sets are shown. Each tier is trained starting with the best model of its previous tier. The learning is generally rational (scores go up) but is less effective in tiers 2 and 3.}
  \label{fig:curricgoroom}
\end{figure}

We initially only train with tier-1 training data. After convergence we then use the best model selected from a dev set of games to initialize the model of tier-2, and so on. Because tiers 1--3 differ significantly from tiers 4--6 (the latter have movements and more games per tier), we alter our approach slightly as training proceeds. 

We start training tier-1 with the games of tier-1 only. When we train tier-2, we mix the games of tier-1 and tier-2 in order to make the agent perform well on both tiers. We then mix tier-3 data in. But for tier-4 to tier-6, we only use the data for the specific stage of training, and do not mix in data from previous tiers. As a final stage, we fine-tune the agent with all training games starting from the best model of the last tier. For each stage of the curriculum learning we initialize  $\epsilon$ to 1 and decay evenly to $1\mathrm{e}{-4}$ across a maximum of two million steps. In ablation experiments without curriculum learning (`mix go-room') we instead decay over 10 million steps.


Training graphs for `curric' with the `go-room' strategy broken down by tier are shown in Figure~\ref{fig:curricgoroom}. For tier-1 we converge to around 95\% of total score after 140 epochs, which means our agent grasps basic cooking abilities. However, the results of tier-2 and tier-3 are flat, indicating there is minor ingredient confusion but it is never resolved. For tiers 4 through 6, scores generally improve from 40\% to roughly 60\%, indicating progressive ability to learn to navigate rooms. Here we can see that a) curriculum training is generally helpful at every tier, and that b) the ability to reach full scores generally decreases by tier. 

Table~\ref{tbl:all-vs-curriculum} breaks down the test results `mixed go-room' and `curric go-room' by tier, evaluating after all training is complete. Curriculum learning generally outperforms mixed training in each tier.

\subsection{Analysis of Universal Information Conversion}
\label{sec:go-strategy-experiments}

We create a simple knowledge graph to collect the room information for each game during both training and inference. When there is a game with unknown room information, we collect this information by random walking with {\it go} and {\it open} actions (e.g. open a sliding door) that are essential for walking through each game. To avoid incurring extra steps at inference time for building the knowledge graph, we take the first 50 steps out of the total 100 available steps to perform the random walk when the floor map has not been collected for a room, and the remaining steps to perform normal game playing for all episodes of each game. The knowledge graph provides extra room information for each {\it go} action, and can be used even when partially built.

Table~\ref{tbl:go-strategy-tier} breaks down the performance of each strategy for dealing with instance information in each relevant tier. The model used for evaluation is the best model on the dev set after the tier-6 training of curriculum learning, without the final fine-tune. It is clear that `go-cardinal,' which does not convert any instance information, is less able to learn than the other methods at any tier. As the number of rooms to navigate grows from tier-4 to tier-6, the random navigation strategy becomes less effective, such that the `go-room' transferring from instance-level cardinal information into universal-level room transition information is the most effective at navigating the large twelve-room games of tier-6.

\begin{table}[ht]
\centering
\begin{tabular}{lc|rrr}
Test                    & Tier & go-cardinal & go-random & go-room \\\hline
\multirow{3}{*}{Test-1} & 4    & 49          & {\bf 58}  & 56      \\
                        & 5    & 40          & 48        & {\bf 49}\\
                        & 6    & 36          & 44        & {\bf 47}\\\hline
\multirow{3}{*}{Test-2} & 4    & 55          & 58        & {\bf 58}      \\
                        & 5    & 50          & 55        & {\bf 60}      \\
                        & 6    & 28          & 41        & {\bf 45}     \\\hline
\end{tabular}
\caption{Breakdown of information conversion strategies of tier-4 to tier-6 on both Test 1 and Test 2 with the best model after curriculumn learning on tier-6, without the final fine-tuning; `go-cardinal' is the worst since the instance-knowledge learning is overfit on each single game. The `go-random' approach is less effective as map size increases, while the `go-room' method performs the best for both test sets.}
\label{tbl:go-strategy-tier}
\end{table}

Table~\ref{tbl:go-strategy-fit} shows that there is a correlation between the most recently trained tier and performance on test data from that tier; we run `curric go-room' but stop after the tier indicated, then subdivide test data per-tier. We see strongest performance on the main diagonal. This is reasonable because the six-room games of tier-4 use the \textbf{same} six rooms each time and so on; the extra rooms of tier-6 aren't known during tier-4 training, and some decay of tier-4 rooms is observed as learning is rededicated to new rooms. Nevertheless, by fine-tuning on all tiers we get the best overall performance.

\begin{table}[!htb]
\centering
\begin{tabular}{l|rrr|r}

\diagbox{Test}{Train} & Tier-4 & Tier-5 & Tier-6 & {\it All} \\ \hline

Tier-4  &  \textbf{62} & 59 &  56 & {\it 68} \\
Tier-5  & 41 & \textbf{50} & 49 & {\it 64} \\
Tier-6 &  26 &  35 &  {\bf 47} & {\it 46} \\
\end{tabular}
\caption{Recency effect of curriculum learning (using go-room) on Test 1; performance on tier-specific subsets is best on the last tier used for training. After fine-tuning on all tiers (the {\it All} column) we get the best overall performance.}
\label{tbl:go-strategy-fit}
\end{table}

\subsection{Improvement from Uncertain Exploration}    
\label{sec:uncertainty-experiment}


\begin{table}[]
\centering
\setlength\tabcolsep{5pt} 
\begin{tabular}{c|rrr|rrr}
\multirow{2}{*}{Tier}    & \multicolumn{3}{c|}{Test 1} & \multicolumn{3}{c}{Test 2} \\
    & \eps & \sampl & UCB & \eps & \sampl & UCB \\ \hline
1   &  96$\pm$2 & 97$\pm$1 & {\bf 100}$\pm$0 & 97$\pm$1 & 100$\pm$0 & {\bf 100}$\pm$0  \\ 
2   &  71$\pm$2 & 75$\pm$0 & {\bf 75}$\pm$0  & {\bf 68}$\pm$2 & 65$\pm$1 & 67$\pm$0        \\ 
3   &  58$\pm$1 & 62$\pm$1 & {\bf 64}$\pm$0  & 67$\pm$3 & 70$\pm$1 & {\bf 71}$\pm$0  \\ 
4   &  67$\pm$3 & 74$\pm$2 & {\bf 77}$\pm$1  & 65$\pm$1 & 69$\pm$2 & {\bf 69}$\pm$0  \\ 
5   &  62$\pm$2 & {\bf 70}$\pm$2 & 69$\pm$0  & 60$\pm$2 & 66$\pm$2 & {\bf 70}$\pm$1  \\ 
6   &  47$\pm$2 & 53$\pm$2 & {\bf 64}$\pm$0  & 47$\pm$2 & {\bf 52}$\pm$1 & 51$\pm$3  \\ \hline
All &  63$\pm$1 & 69$\pm$1 & {\bf 72}$\pm$0  & 63$\pm$1 & 67$\pm$1 & {\bf 68}$\pm$1  \\ 
\end{tabular}
\caption{We compare the evaluation results of {\bf \eps}: using small randomization with $\epsilon=0.05$; {\bf \sampl}: sampling from policy distribution with temperature $T=0.01$; and {\bf UCB}: using LinUCB. LinUCB shows the best results on both test sets. We report the percentage of scores with a confidence level of 0.95 over 10 episodes.}
\label{tbl:curriculum-vs-ft-vs-UCB}
\end{table}

At inference time, we compare different methods of using learned policies. First, as a baseline, we set $\epsilon=0$ to use the learned policy without randomization; this is shown in Table \ref{tbl:all-vs-curriculum} (curric). Then, we add $\epsilon=0.05$ to allow a small stochasticity. We consider sampling from the policy distribution as the third method, by treating Q-values as logits and sampling directly from them by the Gumbel-max sampler \cite{vieira2014gumbel}. To increase the sampling accuracy, we use temperature $T=0.01$ on logits to avoid random selection due to a flat distribution \cite{44873}. Finally, we use the LinUCB method, as discussed in Section \ref{sec:uncertainty}. Since raw scores in our games are either 0 or 1, the learned Q-values will fall in a small range around 1. To avoid large bounds affecting Q-values too much, we normalize bounds at each step for all actions according to the largest bound and use a coefficient of 0.2.

We show the evaluation results in Table \ref{tbl:curriculum-vs-ft-vs-UCB} with a confidence level of 0.95, computed over 10 episodes of game playing for each games in test sets.
Using $\epsilon=0.05$ to add stochasticity shows worse results than the greedy method with $\epsilon=0$ (Table \ref{tbl:all-vs-curriculum}), because randomly choosing a dangerous action could lead to direct failure in these games, as discussed in Section \ref{sec:uncertainty}. Sampling from the policy distribution works much better than these two methods, but is still outperformed by LinUCB on most tiers and the final overall results.

\section{Related Work}
Many recent works \cite{D15-1001,P16-1153,journals/corr/LiMRGGJ16,DBLP:journals/corr/abs-1805-07274,fulda2017affordance,textworld-a-learning-environment-for-text-based-games,DBLP:conf/cig/KostkaKKR17} on building agents of text-based games apply the DQN \cite{google-atari} from playing video games or its variants. Different aspects of DQN have been presented, such as action reduction with language correlation \cite{fulda2017affordance}, a bounding method \cite{DBLP:conf/nips/ZahavyHMMM18}, the introduction of a knowledge graph \cite{DBLP:journals/corr/abs-1812-01628}, text understanding with dependency parsing \cite{my-own} and an entity relation graph \cite{DBLP:journals/corr/abs-1812-01628}.
However, previous work is chiefly focused on learning to self-train on games and then do well on the same games, instead of playing unseen games. \newcite{DBLP:journals/corr/abs-1806-11525} work on generalization of agents on variants of a very simple coin-collecting game. The simplicity of their games enables them to use an LSTM-DQN method with a counting-based reward. \newcite{DBLP:journals/corr/abs-1812-01628} use a knowledge graph as a persistent memory to encode states, while we use a knowledge graph to make actions more informative.

\section{Conclusion}

In this paper, we train agents to play a family of text-based games. Instead of repeatedly optimizing on a single game, we train agents to play familiar but unseen games. 
We show that curriculum learning helps the agent learn better. We convert instance knowledge into universal knowledge via map familiarization. We also show how the incorporation of bandit feedback to both training and evaluation phases leads the agent to explore more thoroughly and reach higher scores.

\bibliographystyle{named}
\bibliography{ijcai20}

\end{document}